\documentclass[sigconf]{acmart}
\usepackage{multirow}
\AtBeginDocument{%
  }
\usepackage{CJKutf8}

\copyrightyear{2026}
\acmYear{2026}
\setcopyright{cc}
\setcctype{by}
\acmConference[MM '26]{Proceedings of the 34th ACM International Conference on Multimedia}{November 10--14, 2026}{Rio de Janeiro, Brazil}
\acmBooktitle{Proceedings of the 34th ACM International Conference on Multimedia (MM '26), November 10--14, 2026, Rio de Janeiro, Brazil}
\acmDOI{10.1145/3767308.3835569}
\acmISBN{979-8-4007-2213-4/2026/11}

\makeatletter

\newcommand{\makesupplementtitle}{%
  \clearpage
  \begingroup

    \let\@vspace\@vspace@orig
    \let\@vspacer\@vspacer@orig
    \let\@footnotemark\@footnotemark@nolink
    \let\@footnotetext\@footnotetext@nolink

    \renewcommand\thefootnote{\@fnsymbol\c@footnote}%
    \setcounter{footnote}{0}%

    \hsize=\textwidth
    \def\@makefnmark{%
      \hbox{\@textsuperscript{\@thefnmark}}%
    }%

    \@mktitle
    \@mkauthors

    \thispagestyle{firstpagestyle}%
    \noindent
    \twocolumn[\box\mktitle@bx]%

    \setcounter{footnote}{0}%
    \def\@makefnmark{%
      \hbox{\@textsuperscript{\normalfont\@thefnmark}}%
    }%

    \@titlenotes
    \@subtitlenotes
    \@authornotes

  \endgroup

  \setcounter{footnote}{0}%
  \@afterindentfalse
  \@afterheading
}

\makeatother

\begin{document}

\title{DTRNet: Dual Text-Radical Decoding for Handwritten Chinese Text Recognition with Faked Character Detection}

\author{Runrui Li}
\orcid{0000-0001-6861-9887}
\affiliation{%
  \department{School of Artificial Intelligence}
  \institution{Beijing Normal University}
  \institution{Beijing Key Laboratory of Artificial Intelligence for Education}
  \institution{Engineering Research Center of Intelligent Technology and Educational Application}
  \city{Beijing}
  \country{China}}
\email{runruili@mail.bnu.edu.cn}

\author{Lin Zhu}
\orcid{0000-0001-6487-0441}
\correspondingauthor
\affiliation{%
  \department{School of Artificial Intelligence}
  \institution{Beijing Normal University}
  \institution{Beijing Key Laboratory of Artificial Intelligence for Education}
  \institution{Engineering Research Center of Intelligent Technology and Educational Application}
  \city{Beijing}
  \country{China}}
\email{linzhu@bnu.edu.cn}

\author{Hua Huang}
\orcid{0000-0003-2587-1702}
\affiliation{%
  \department{School of Artificial Intelligence}
  \institution{Beijing Normal University}
  \institution{Beijing Key Laboratory of Artificial Intelligence for Education}
  \institution{Engineering Research Center of Intelligent Technology and Educational Application}
  \city{Beijing}
  \country{China}}
\email{huahuang@bnu.edu.cn}

\renewcommand{\shortauthors}{Runrui Li, Lin Zhu, and Hua Huang}

\begin{abstract}

In K-12 educational scenarios, handwritten Chinese text recognition should not only transcribe student writing, but also detect faked characters. However, existing recognition models are usually confined to a predefined set of normal characters and therefore cannot explicitly identify faked characters. Existing detection methods exhibit complementary limitations: character-level methods provide interpretable structural evidence but suffer from low efficiency, whereas line-level methods are efficient but rely heavily on confidence scores, making them prone to missed detections and lacking explicit structural evidence. Thus, the key challenge is to preserve character-structural evidence independent of contextual inference while maintaining line-level efficiency. To this end, we propose DTRNet, a dual Text-Radical decoding framework for line-level faked character detection. DTRNet decouples context-aware text recognition from character-wise structural verification, where the text branch performs line-level transcription and the radical branch predicts legal Ideographic Description Sequences (IDS) for lexicon-based faked character judgment. We further introduce IDS-Guided Confidence Adjustment (IGCA) to refine text predictions using structural evidence during inference. Experimental results demonstrate that DTRNet effectively detects faked characters while maintaining strong recognition performance and providing interpretable radical-level evidence. Code, checkpoints, and the processed dataset are publicly available at \url{https://github.com/BNU-ERC-ITEA/DTRNet}.
\end{abstract}




\begin{teaserfigure}
  \includegraphics[width=0.68\textwidth]{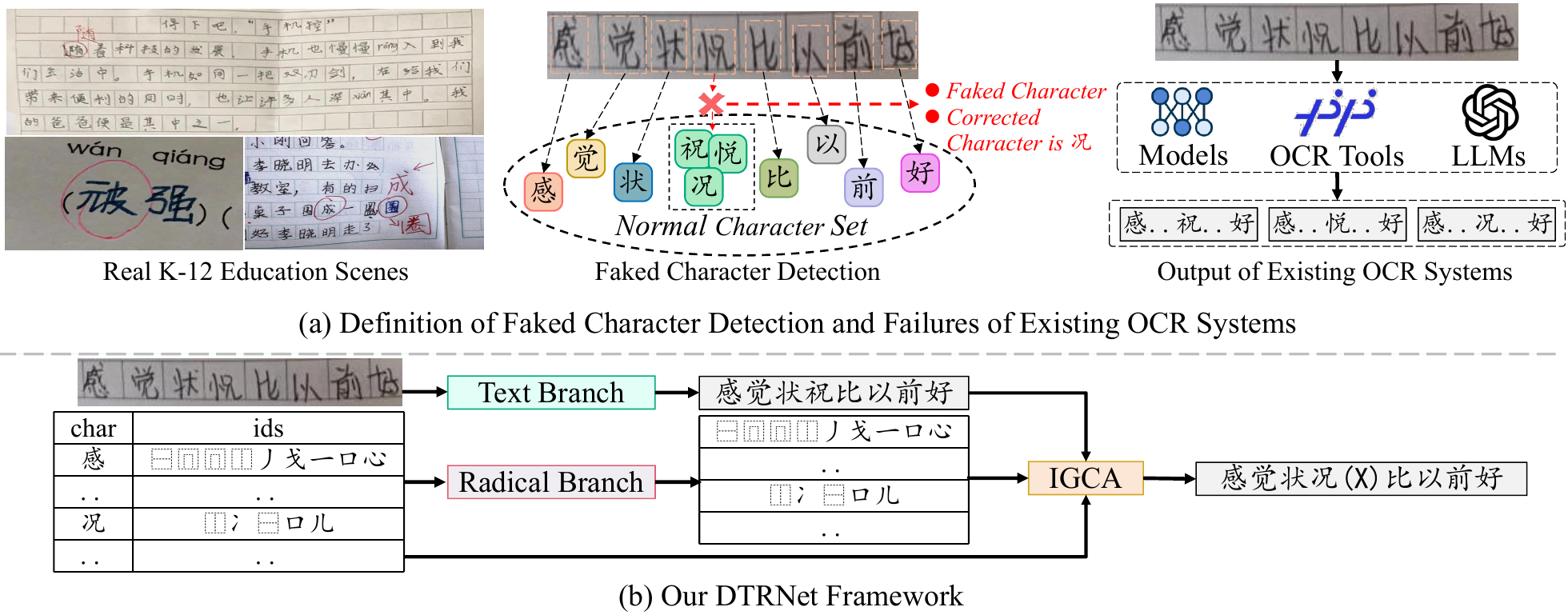}
  \centering
    \caption{Overview of faked character detection and the proposed DTRNet framework. (a) Real K-12 education examples. In Chinese writing assessment and automatic grading, teachers need to identify faked characters in student handwriting so that original writing errors can be marked rather than overlooked. We further illustrate the definition of faked character detection and typical failure cases in existing recognition systems. (b) Architecture of DTRNet, which combines text and IDS (Ideographic Description Sequence) branches with IGCA (IDS-Guided Confidence Adjustment) to improve faked character-aware recognition.}
    \Description{A two-part overview. The upper panel shows K-12 handwriting examples, a malformed handwritten character outside the normal character set, and existing OCR systems that normalize it into plausible characters. The lower panel sends a text line through parallel text and radical-IDS branches, whose outputs are combined by IGCA to mark the faked character as X.}
  \label{fig:intro}
\end{teaserfigure}

\maketitle 

\section{Introduction}

Handwritten Chinese text recognition (HCTR) has achieved remarkable progress in recent years \cite{scut-hccdoc_2020,casia_hwdb}. It is widely used in K-12 education, such as Chinese writing assessment and automatic grading \cite{IGS_3,IGS_2,jiangseeing}. However, handwritten text in these scenarios is not always composed of normal characters from standard dictionaries. Due to writing mistakes, memory lapses, or radical confusion, students may produce faked characters \cite{visual_c3}, namely non-existent characters that do not correspond to any normal character in standard dictionaries. Because such characters are often visually similar to normal ones, existing recognition models tend to over-correct them into plausible normal characters. This not only obscures the original writing errors, but also undermines the reliability of feedback and transcription. Therefore, HCTR systems for these scenarios should not only accurately transcribe normal text, but also detect faked characters at the text-line level.

Existing studies still exhibit clear limitations in handling faked characters, as shown in Fig.~\ref{fig:intro}. Existing OCR systems \cite{crnn_2017,paddleocr_2020,gpt} tend to over-correct faked characters into plausible normal characters under contextual priors, while NLP-based Chinese Spelling Correction (CSC) \cite{NLP_CSC1,NLP_CSC2} methods operate on text rather than images and therefore cannot capture the visual structures of faked characters. More dedicated methods mainly follow two paradigms. Character-level Handwritten Chinese Character Error Correction (HCCEC) \cite{TAN,CDC,hccec_bmvc} methods exploit structural cues such as radicals. While interpretable, they first segment text lines into isolated characters and then detect them one by one, resulting in low efficiency in text-line scenarios. Line-level methods \cite{visual_c3,wang2025viscgec} perform text-line recognition and detect suspicious characters through confidence-based filtering. Although more efficient, their reliance on confidence scores often causes missed detections, makes it difficult to distinguish ordinary recognition errors from faked characters, and provides no explicit structural evidence. Meanwhile, in general text recognition, line-level recognition has become the mainstream pipeline because of its advantages in both efficiency and accuracy. Therefore, introducing explicit character-structure modeling into line-level faked character detection is essential for improving detection capability and interpretability while preserving the efficiency of text-line processing.

To address this issue, we propose DTRNet, a dual Text-Radical decoding framework for HCTR with faked character detection. The key idea is to explicitly decouple context-aware text recognition from character-wise structural verification, so that faked character detection is not determined solely by recognition confidence. Specifically, the text branch performs line-level transcription to preserve normal text recognition performance, while the Radical branch extracts local structural representations at each character position and transcribes them into Ideographic Description Sequence (IDS) \cite{zhang2018radical,zhang2020radical}, providing structural evidence independent of contextual inference. Based on this design, faked characters are identified through character-wise structural verification against a predefined Character-IDS lexicon. Furthermore, we introduce IDS-Guided Confidence Adjustment (IGCA), which feeds structural evidence back to the text branch to calibrate context-driven predictions. In this way, DTRNet unifies text transcription, structural verification, and faked character detection within a single line-level framework. 
Since faked character categories are inherently open-set and cannot be exhaustively enumerated in advance, we reconstruct Visual-C3 \cite{visual_c3} as a generalized zero-shot \cite{GZSL_1,GZSL_2} benchmark for line-level faked character detection. Experiments show that DTRNet consistently outperforms representative line-level baselines in both text-line recognition and faked character detection, while providing interpretable radical-level evidence. Comparisons with OCR tools and MLLMs further confirm its effectiveness for faked character detection. Moreover, results on the public BCTR benchmark \cite{BCTR} demonstrate DTRNet's robustness and generalization.

In summary, the contributions of this paper are:
\begin{itemize}
\item We propose DTRNet, a dual Text-Radical decoding framework that adds explicit character-structure modeling to HCTR for accurate and interpretable faked character detection.

\item We design a legality-aware IDS decoding mechanism for the radical branch and further propose IDS-Guided Confidence Adjustment (IGCA), which together improve the reliability of structural verification and the robustness of text recognition.

\item We reconstruct Visual-C3 as a generalized zero-shot benchmark for line-level faked character detection. Experimental results demonstrate that DTRNet consistently outperforms representative baselines, OCR tools, and MLLMs on this task, while also exhibiting strong robustness and generalization on the public BCTR benchmark.

\end{itemize}

\section{Related Work}
\begin{figure}[h]
    \centering
   \includegraphics[width=0.6\linewidth]{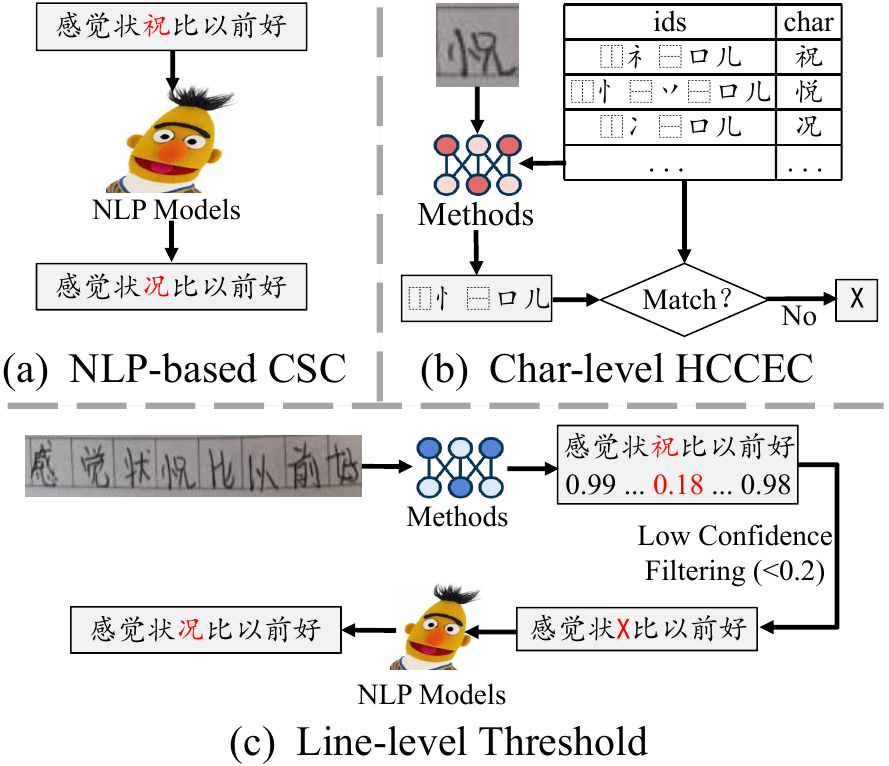}
    \caption{Representative paradigms in existing studies for handling faked characters. (a) NLP-based Chinese Spelling Correction (CSC) methods operate on text-only inputs and therefore cannot model visually written faked characters. (b) Character-level Handwritten Chinese Character Error Correction (HCCEC) methods exploit structural cues such as radicals to detect faked characters, but require isolated character processing. (c) Line-level methods detect suspicious characters through low-confidence filtering after text-line recognition, which lacks explicit structural evidence.}
    \Description{Three paradigms for faked character processing: text-only Chinese spelling correction, character-level IDS prediction followed by lexicon matching, and line-level recognition followed by low-confidence filtering and language correction.}
    \label{fig:related_fake}
\end{figure}

This section reviews four lines of research related to our work: Handwritten Chinese Text Recognition (HCTR), Chinese Spelling Correction (CSC), Handwritten Chinese Character Error Correction (HCCEC), and line-level Chinese faked character detection. In this paper, ``faked characters'' refer to invalid characters caused by writing errors that are absent from standard dictionaries and encoding systems. As shown in Fig.~\ref{fig:related_fake}, existing studies mainly fall into three paradigms: text-only CSC, character-level HCCEC, and line-level threshold-based detection.

\noindent\textbf{Handwritten Chinese Text Recognition.}
HTR has evolved from CNN/RNN-based models to attention-based encoder-decoder frameworks \cite{vaswani2017attention}. CRNN \cite{crnn_2017} established a unified framework for feature extraction, sequence modeling, and transcription. NRTR \cite{nrtr_2019} replaced convolution and recurrence with a pure Transformer architecture. ABINet \cite{abinet_2021} introduced explicit language modeling to better exploit contextual information, while SVTR \cite{svtr_2021} improved recognition through stronger visual modeling alone. More recently, multimodal large language models \cite{gpt,Qwen-VL,Qwen2-VL,Qwen2.5-VL,team2026kimi,bytedance_doubao2024} and OCR-oriented large models \cite{paddleocr_2020, gotocr, Mistral_OCR, li2025monkeyocrdocumentparsingstructurerecognitionrelation} have extended OCR from text recognition to broader document understanding tasks \cite{OCRBench_2024,fu2024ocrbenchv2improvedbenchmark}. 

Despite their strong recognition ability, these methods are designed to recover readable text rather than explicitly detect faked characters. As stronger linguistic priors are introduced, models tend to over-correct abnormal inputs into plausible normal characters, which benefits transcription but suppresses faked character detection. As a result, existing HCTR methods lack explicit and interpretable evidence for identifying faked characters.

\noindent\textbf{Chinese Spelling Correction.}
CSC is a classical NLP task for detecting and correcting erroneous characters in text. Representative methods have evolved from heuristic designs to pretrained language model frameworks. Soft-Masked BERT \cite{soft_bert} connects explicit error detection with correction through soft masking. SpellGCN \cite{spellgcn} injects phonetic and visual confusion relationships into language models via graph convolution. PLOME \cite{liu2021plome} further incorporates pinyin and structural information during task-specific pretraining.

In essence, CSC methods are text-only, as shown in Fig.~\ref{fig:related_fake}(a). Benchmarks such as SIGHAN13/14/15 \cite{NLP_CSC1,NLP_CSC3,NLP_CSC4} represent errors purely in text form and thus only cover characters defined in standard encoding systems. Consequently, CSC can address misspellings among valid characters, but cannot model visually written faked characters in handwriting images.

\noindent\textbf{Handwritten Chinese Character Error Correction.}
HCCEC has recently emerged as a dedicated task for real visual writing errors. Since faked character categories are open-ended, conventional closed-set classification is inadequate. TAN \cite{TAN} formulates HCCEC as a generalized zero-shot learning (GZSL) problem \cite{GZSL_1} and explicitly models internal character layouts through tree-structured decoding. CDC \cite{CDC} further reduces the bias of auto-regressive IDS decomposition to improve generalization to unseen faked characters. Recently, a multimodal method \cite{hccec_bmvc} advances this task through image-IDS alignment. These methods establish structural cues as essential for open-category recognition.

Still, existing HCCEC methods are character-level, as shown in Fig.~\ref{fig:related_fake}(b). Although they provide interpretable structural evidence, they require explicit character segmentation and per-character processing, which limits efficiency in text-line scenarios.

\noindent\textbf{Text-line Level Chinese Faked Detection.}
Text-line faked character detection extends the problem from isolated characters to continuous handwritten text. Visual-C3 \cite{visual_c3} introduces the first Chinese text-line image dataset containing real handwritten faked characters. VisCGEC \cite{wang2025viscgec} further studies visual Chinese grammatical error correction directly from handwritten sentence images, covering faked characters, misspellings, and broader grammatical errors. These works provide valuable data resources and task settings for studying real visual writing errors.

However, existing line-level methods still have clear limitations. As shown in Fig.~\ref{fig:related_fake}(c), Visual-C3 mainly detects suspicious characters through low-confidence filtering after text-line recognition. Although efficient, this strategy relies on posterior confidence rather than explicit character structure, making it difficult to distinguish ordinary recognition errors from genuine faked characters and providing no interpretable structural evidence. Moreover, its random data split may expose some faked characters during training, weakening the evaluation of generalization to unseen cases. Therefore, an efficient line-level framework with explicit character-structure modeling is still lacking.

Overall, prior studies suggest that effective faked character detection requires both line-level efficiency and explicit structural evidence. This motivates explicit character-structure modeling in a text-line framework for improved detection and interpretability.

\section{Proposed Method}
\subsection{Motivation and Task Definition}
\label{ssec:method_motivation}

Existing HCTR models are primarily designed to recognize normal character sequences from text-line images. To improve accuracy, they rely on contextual language priors during decoding. However, this semantic dependence inevitably induces an over-correction effect: it tends to force faked characters into plausible normal ones, thereby obscuring actual student writing errors in K-12 educational settings. Therefore, line-level faked character detection cannot rely solely on the recognized text sequence, but also requires intra-character structural evidence independent of contextual inference.

HCCEC studies show that faked character categories are essentially infinite. \textit{Inspired by this, we formulate line-level faked character detection as GZSL problem: models trained solely on text lines composed of normal characters, yet they must accurately transcribe normal characters while explicitly rejecting unseen faked ones during inference.} Unlike character-level HCCEC, the line-level setting introduces an additional challenge: contextual priors help recognize poorly written characters but also cause faked characters to be misrecognized as normal ones.

\begin{figure*}[ht]
    \centering 
    \includegraphics[width=0.68\textwidth]{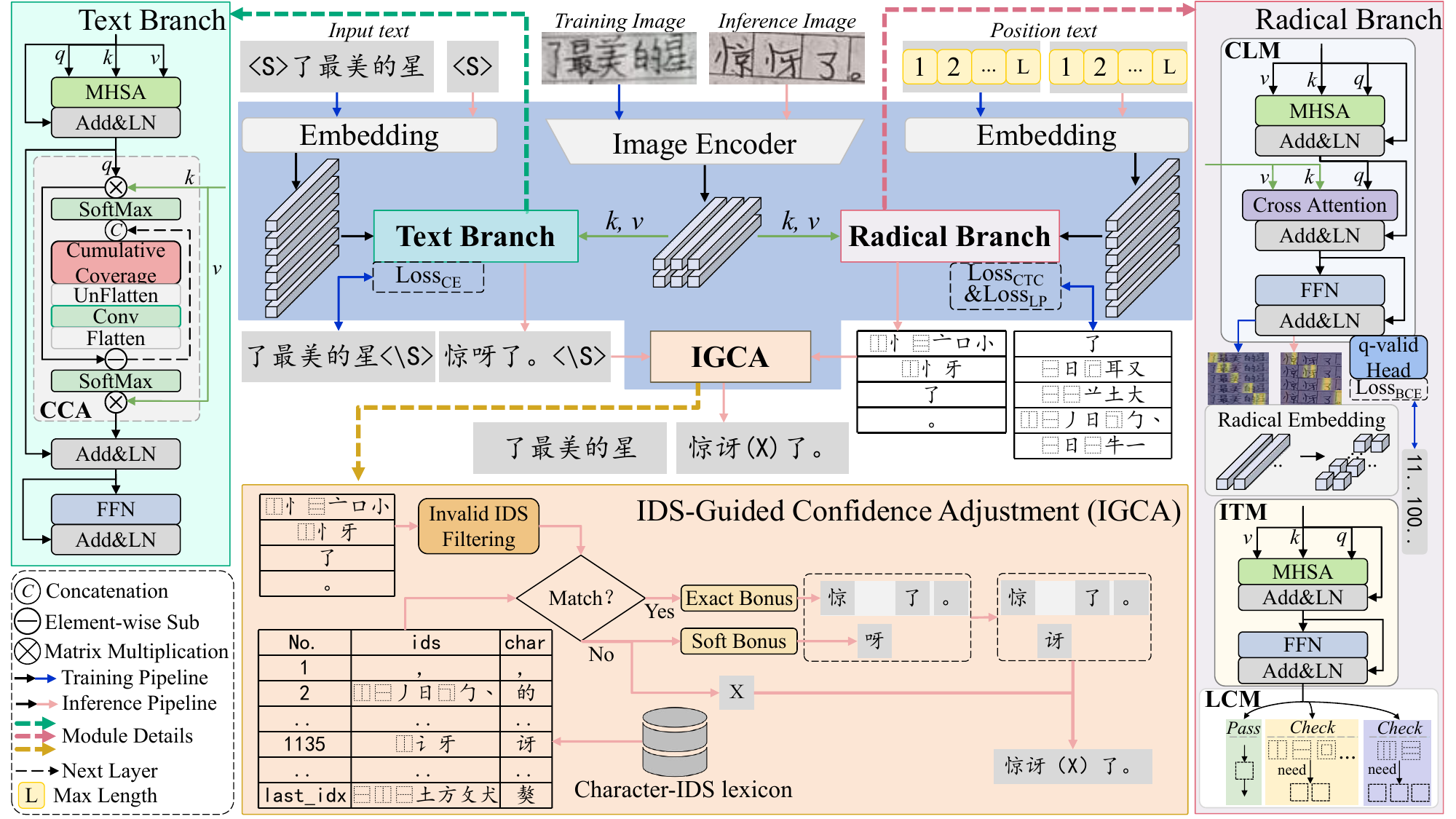} 
    \caption{Overview of the proposed DTRNet framework. Given an input text-line image, a visual encoder (Sec.~\ref{ssec:method_Encoder}) extracts feature map $F$, which is fed into two decoupled decoding branches. The text branch performs context-aware transcription to recognize normal texts (Sec.~\ref{ssec:method_textDecoder}), while the radical branch (Sec.~\ref{ssec:method_radicalDecoder}) conducts character-level structural modeling, including character localization, IDS transcription, and legality constraints, to produce structure-aware representations independent of contextual inference. During training, both branches are jointly optimized with additional legality supervision on IDS sequences. During inference, predicted IDS sequences are matched with a pre-defined Character-IDS lexicon to identify faked characters, and IDS-Guided Confidence Adjustment (IGCA, Sec.~\ref{ssec:method_igca}) is applied to calibrate text branch predictions. This design enables unified modeling of text recognition, structural verification, and faked character detection.} 
    \Description{The DTRNet training and inference pipeline. An image encoder feeds separate text and radical branches. The radical branch localizes characters, predicts IDS sequences under legality constraints, and supplies structural evidence to IGCA, which refines the text output or assigns the faked-character label X.}
    \label{fig:framework}
\end{figure*}

Motivated by these observations, we propose DTRNet, a dual-branch Text–Radical decoding framework for HCTR and faked character detection, as shown in Fig.~\ref{fig:framework}. It detects faked characters through IDS-based matching and further improves recognition with IGCA during inference.

\subsection{Encoder Architecture}
\label{ssec:method_Encoder}

To preserve both fine-grained intra-character structures and line-level context, we adopt the SVTR encoder \cite{svtr_2021} as the visual backbone of DTRNet. Its local mixing blocks capture stroke- and radical-level patterns, while its global mixing blocks model long-range dependencies, yielding visual features suitable for both text transcription and structural verification.

For an input image $X \in \mathbb{R}^{H \times W \times 3}$, patch embedding transforms the image into a sequence with dimensions $\frac{H}{4} \times \frac{W}{4} \times C_0$. The backbone comprises three stages for feature extraction, where each stage contains six local or global mixing blocks. After the second stage, we apply downsampling to halve the feature height and increase the channel dimension while keeping the width unchanged. The encoder finally outputs the feature map $F \in \mathbb{R}^{\frac{H}{8} \times \frac{W}{4} \times C}$.

\subsection{Text Decoder}
\label{ssec:method_textDecoder}

The text branch is based on a Transformer decoder and transcribes the shared visual feature map $F$ into the target text sequence $Y_{\text{text}}$. As shown in the left panel of Fig.~\ref{fig:framework}, we first flatten $F \in \mathbb{R}^{\frac{H}{8} \times \frac{W}{4} \times C}$ into a visual sequence of length $N = \frac{H}{8} \times \frac{W}{4}$. To reduce repeated attention to the same region, we introduce Cross-Coverage Attention (CCA) \cite{coverage_atten,zhao2022comer,zhu2025tamer} to model historical alignment.

Let $\alpha_t^i \in \mathbb{R}^{N}$ denote the cross attention distribution at time step $t$ in the $i$-th decoder layer, $E_t^i \in \mathbb{R}^{N}$ denote the corresponding raw attention logits. We first aggregate the historical attention from previous time steps and combine it with the attention from the previous layer to form a coverage term:
\begin{equation}
C_t^i = \mathrm{Flat}\!\left(\mathrm{Conv2D}\!\left(\mathrm{Reshape}\!\left(\sum_{\tau=1}^{t-1}[\alpha_\tau^i;\alpha_\tau^{i-1}]\right)\right)\right),
\end{equation}
where $[\cdot;\cdot]$ denotes channel concatenation, $\mathrm{Reshape}(\cdot)$ restores the 1-D attention sequence to a 2-D spatial map, $\mathrm{Conv2D}(\cdot)$ aggregates local neighborhood information, and $\mathrm{Flat}(\cdot)$ flattens it back to a vector. The refined attention is then computed as
\begin{equation}
\hat{\alpha}_t^i = \mathrm{Softmax}(E_t^i - C_t^i).
\end{equation}
This mechanism accumulates visited regions over time and spreads coverage to neighboring locations. It helps the text decoder avoid repeated attention on the same character region and focus on unexplored regions, which improves transcription accuracy.

\subsection{Radical Decoder}
\label{ssec:method_radicalDecoder}

As shown in the right panel of Fig.~\ref{fig:framework}, the radical branch provides character-structural evidence independent of linguistic context for line-level faked character detection. It contains three components: character localization, IDS transcription, and legality-aware decoding. The first two components establish radical-wise structural representations from text-line features, while the legality-aware decoding mechanism further regularizes IDS prediction to make the resulting structural evidence more reliable for verification. With this design, the radical branch can preserve local character structure while remaining compatible with efficient line-level processing.

\noindent\textbf{Char Localization Module.} Similar to the text branch, we first flatten the encoder output $F$ into a feature sequence. To obtain character-level features, we introduce a set of learnable queries $Q = \{q_1, q_2, \ldots, q_L\}$, where $L$ is the maximum text length. A lightweight CLM based on Transformer decoder executes cross attention between $Q$ and $F$:
\begin{equation}
F^{\mathrm{char}} = \mathcal{D}_{\mathrm{loc}}(Q, F) \in \mathbb{R}^{L \times C} \text{,}
\end{equation}
where $F^{\mathrm{char}}_i \in \mathbb{R}^{C}$ denotes the feature of the $i$-th character. Since $L$ is the maximum text length, some queries are redundant for shorter inputs. We introduce a query validity head to predict valid queries and construct binary labels where only the first $T$ positions are marked as valid. During inference, the predicted validity is transformed into a monotonic prefix mask to infer the number of queries. Invalid queries are discarded before IDS transcription module to ensure clean structural inputs. This design localizes characters without relying on text decoding states, reducing contextual bias and providing clean inputs for structural modeling.

\begin{CJK*}{UTF8}{gbsn}
\begin{figure}[h]
    \centering % 图片居中
    \includegraphics[width=0.63\columnwidth]{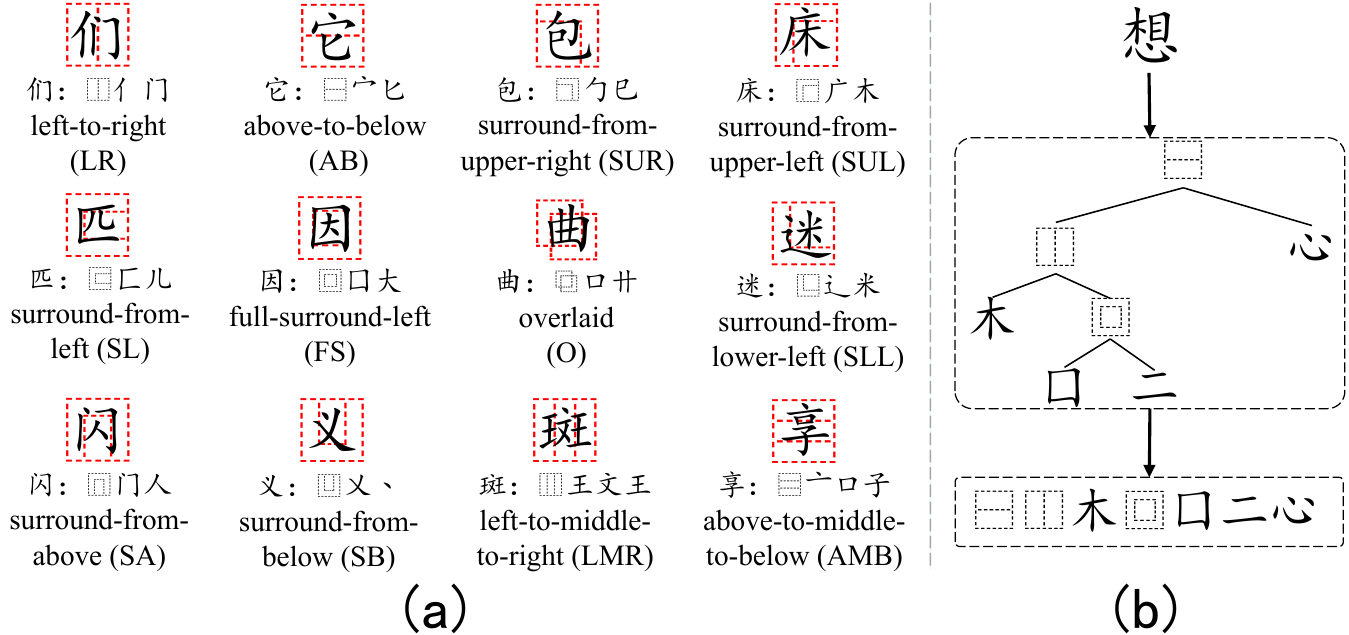}
    \caption{The Ideographic Description Sequence (IDS) representation, including 10 binary structures and 2 trinary structures. (a) Spatial structural tokens and their corresponding arities. (b) The hierarchical tree structure and the flattened 1D IDS token sequence for the Chinese character ``\textmd{想} (xiang)''.}
    \Description{The twelve IDS spatial composition operators with example characters and their arities, followed by the decomposition of the Chinese character xiang into a hierarchical component tree and a flattened one-dimensional IDS token sequence.}
    \label{fig:ids} 
\end{figure}
\end{CJK*}

\noindent\textbf{IDS Transcription Module.} As shown in Fig.~\ref{fig:ids}, the internal structure of a Chinese character can be represented as an Ideographic Description Sequence (IDS), which consists of structural tokens and radical tokens. Structural tokens describe the composition relations among components, while radical tokens correspond to basic character units. Instead of directly predicting character labels, we transcribe each valid character feature into an IDS token sequence to explicitly model its internal structure. Given a character feature $f_i \in \mathbb{R}^{C}$, we first expand it into a sequence of $K$ local frames:
\begin{equation}
U_i = \mathcal{E}_{\mathrm{frame}}(f_i) \in \mathbb{R}^{K \times C},
\end{equation}
where $K$ is a fixed number of frames for each character. $U_i$ is encoded by a Transformer and projected to frame-wise IDS token distributions. We adopt CTC \cite{CTC_2006} instead of autoregressive decoding. CTC performs frame-wise prediction without dependence on past outputs, making it more suitable for faithful transcription. This property helps preserve abnormal but visually plausible structures in faked characters, providing effective evidence for detection.

\noindent\textbf{Legality Constraints Module.} A key challenge of IDS-based structural verification is that visually plausible predictions are not always structurally legal. To address this issue, we design a Legality Constraints Module (LCM) to explicitly inject IDS grammar into decoding. Different from prior validity modeling strategies based on tree decoding or stack-style parsing, our LCM introduces a lightweight Need Counter that tracks the number of pending child nodes during IDS generation, making legality control easy to integrate into both training and inference.

Specifically, let $z_k$ denote the $k$-th predicted token and $n_k$ denote its current need state. We initialize $n_1 = 1$ to indicate the requirement for a root node, and update the state as
\begin{equation}
n_{k+1}=n_k+\mathrm{arity}(z_k)-1,
\end{equation}
where $\mathrm{arity}(z_k)$ is the number of children required by token $z_k$. A legal IDS sequence must satisfy $n_k>0$ throughout decoding and $n_{K+1}=0$ at termination. Based on this formulation, we further impose legality-aware penalties during training and mask illegal candidates during inference, which substantially improves the reliability and stability of IDS prediction.

\subsection{Structural Feedback from Radical Branch}
\label{ssec:method_igca}
Radical branch not only provides structural evidence for faked character detection, but also offers a way to refine context-driven text prediction. To exploit this property, we further propose IDS-Guided Confidence Adjustment (IGCA), which feeds structural evidence back to text branch during inference. In this way, DTRNet uses the radical branch not only for post hoc verification, but also to correct context-induced over-correction in text recognition.

Let $o_t \in \mathbb{R}^{|V|}$ denote the raw logit of the text branch at time step $t$. The adjusted output is defined as
\begin{equation}
\tilde{o}_t = o_t + b_t,
\end{equation}
where $b_t \in \mathbb{R}^{|V|}$ is a bonus vector constructed from the radical branch result. IGCA is activated only when the predicted IDS is complete and structurally legal. If it exactly matches an entry in the predefined Character-IDS lexicon, an exact bonus is added to the matched character class. Otherwise, a soft bonus is assigned to the most similar lexicon entry according to normalized edit distance.

Different from confidence-threshold heuristics that use low confidence as an indirect abnormality signal, IGCA calibrates text predictions with explicit structural evidence. This design preserves the recognition strength of the text branch while improving robustness against context-induced over-correction.

\begin{table*}[h]
\centering
\small
\renewcommand{\arraystretch}{0.4}
\setlength{\tabcolsep}{9pt}
\caption{
Comparison with representative methods on the reconstructed line-level faked character detection benchmark. Text recognition performance on \textit{test\_correct} and \textit{test\_faked} is reported with ACC and 1-NED, and faked character detection is evaluated with Precision (P), Recall (R), and F1. IDS legality denotes the percentage of predicted legal IDS outputs.}
\label{tab:comparison_main}
\begin{tabular}{c|cc|cc|ccc|c|c}
\toprule
\multirow{2}{*}{Method (Venue)} 
& \multicolumn{2}{c|}{\textbf{test\_correct}} 
& \multicolumn{2}{c|}{\textbf{test\_faked}} 
& \multicolumn{3}{c|}{\textbf{Faked Char Detection}} 
& \multirow{2}{*}{\textbf{IDS legality}} 
& \multirow{2}{*}{\textbf{Params} (M)} \\
& \textbf{ACC} & \textbf{1-NED} 
& \textbf{ACC} & \textbf{1-NED} 
& \textbf{P} & \textbf{R} & \textbf{F1} 
&  &  \\
\midrule
CRNN (TPAMI 2016)        & 76.76 & 93.52 &  1.69 & 73.75 & 45.51 &  4.06 &  7.46 & --    & 17.69 \\
NRTR (ICDAR 2019)        & 60.32 & 86.57 &  7.20 & 66.39 & 23.39 & 44.13 & 30.57 & --    & 47.25 \\
ABINet (CVPR 2021)       & 83.21 & 95.70 &  0.50 & 75.15 & 53.19 &  0.97 &  1.91 & --    & 42.86 \\
SVTR (IJCAI 2021)        & 83.16 & 95.74 &  0.13 & 74.80 & 48.57 &  0.33 &  0.66 & --    & 18.87 \\
PARSeq (ECCV 2022)       & 82.72 & 95.76 &  0.18 & 74.21 & 38.46 &  0.58 &  1.15 & --    & 19.26 \\
LISTER (ICCV 2023)       & 82.36 & 95.59 &  0.11 & 74.54 & 52.94 &  0.35 &  0.69 & --    & 21.57 \\
SMTR (AAAI 2025)         & \underline{84.27} & \underline{96.20} &  0.80 & 74.99 & 56.49 &  1.70 &  3.30 & --    & 18.06 \\
CPPD (TPAMI 2025)        & 81.24 & 95.27 & \underline{11.48} & \underline{77.47} & \underline{45.59} & \underline{22.25} & \underline{29.90} & --    & 29.27 \\
SVTRv2 (ICCV 2025)       & 84.02 & 96.09 &  0.04 & 74.94 & \textbf{66.67} &  0.15 &  0.31 & --    & 20.94 \\
\midrule
\textbf{DTRNet}          & \textbf{86.81} & \textbf{96.76} & \textbf{20.29} & \textbf{80.25} & 45.05 & \textbf{34.53} & \textbf{39.10} & \textbf{87.03} & 27.59 \\
\bottomrule
\end{tabular}
\end{table*}

\subsection{Training and Inference}
We train DTRNet end-to-end by jointly optimizing the text branch and the radical branch. Given a training sample $(X, Y_{\text{text}}, Y_{\text{ids}})$, the overall objective is
\begin{equation}
\mathcal{L} = \lambda_{\text{text}} \mathcal{L}^{\text{text}}_{\text{CE}} + \lambda_{\text{ids}} \mathcal{L}^{\text{ids}}_{\text{CTC}} + \lambda_{\text{valid}} \mathcal{L}^{\text{valid}}_{\text{BCE}} + \lambda_{\text{illegal}} \mathcal{L}_{\text{LP}},
\end{equation}
where $\mathcal{L}^{\text{text}}_{\text{CE}}$ is the cross-entropy loss of the text branch, $\mathcal{L}^{\text{ids}}_{\text{CTC}}$ is the character-wise CTC loss of the radical branch, $\mathcal{L}^{\text{valid}}_{\text{BCE}}$ is the binary classification loss for the query validity head, and $\mathcal{L}_{\text{LP}}$ is the legality penalty for IDS prediction.

During inference, the text branch and radical branch operate in parallel. The radical branch first uses the query validity head to determine valid character queries, and then decodes the corresponding character features into IDS sequences under legality constraints. A character is recognized as normal only if its predicted IDS is both structurally legal and matched in the predefined Character-IDS lexicon; otherwise, it is regarded as a faked character. Following Visual-C3, we use the special symbol $X$ to denote faked characters. Afterward, IGCA is applied to calibrate the text branch predictions using the structural evidence from the radical branch. In this way, DTRNet unifies text recognition, structural verification, and faked character detection in a single inference pipeline.

\begin{figure*}[h]
    \centering
    \includegraphics[width=0.65\textwidth]{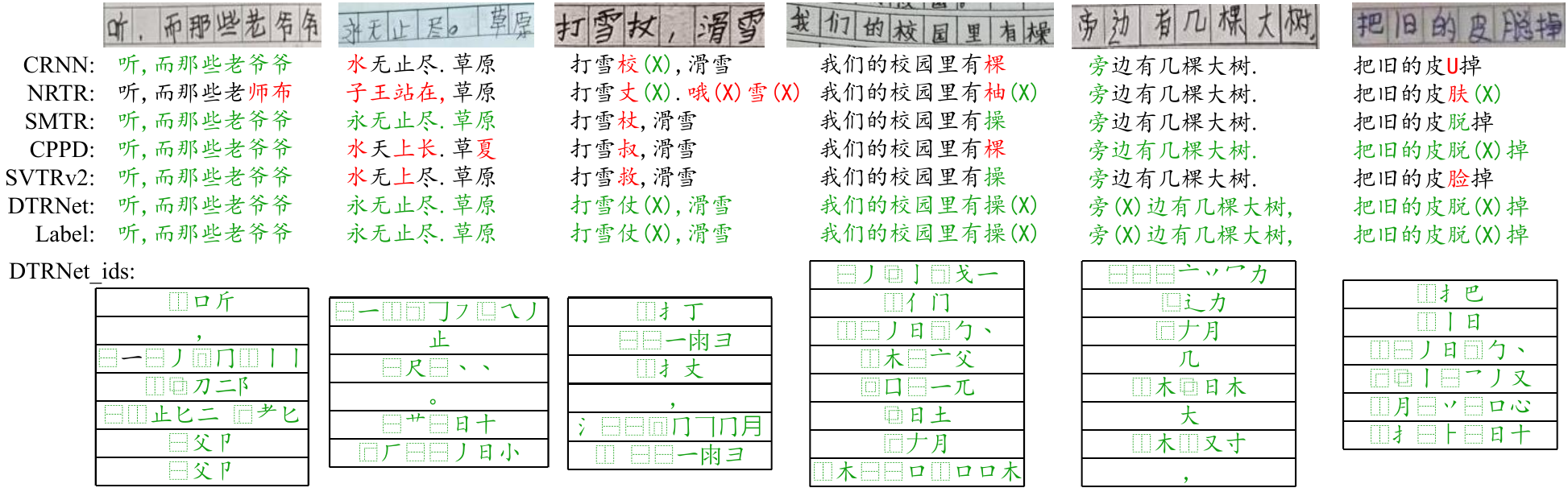} 
    \caption{Qualitative comparison on representative samples from \textit{test\_correct} and \textit{test\_faked}. Green text indicates correct outputs and red text indicates errors. DTRNet maintains competitive recognition on normal samples and detects faked characters as $X$ with character-wise IDS evidence.}
    \Description{Six handwritten text-line examples comparing CRNN, NRTR, SMTR, CPPD, SVTRv2, DTRNet, and ground truth. Correct outputs are green and errors are red. Character-wise IDS predictions show how DTRNet identifies faked characters as X while retaining structural evidence.}
    \label{fig:experiment_compare} 
\end{figure*}

\section{Experiment}
\subsection{Dataset Construction}
Faked characters form an open-set problem, so the original random split of Visual-C3 is not suitable for evaluating zero-shot generalization. We therefore reconstruct it as a line-level GZSL benchmark: the training set contains only normal text lines, while the test set contains both normal lines and lines with unseen faked characters. Using the public train and validation splits (8,058 samples), we crop multi-line images into single lines, yielding 49,337 training samples, 5,482 samples in \textit{correct}, and 4,370 samples in \textit{faked}; all 5,119 faked characters are retained in \textit{faked} for zero-shot evaluation.

We further build a Character-IDS lexicon with 3,020 characters, 323 radical tokens, and 12 structural tokens. During inference, a character is flagged as faked when its predicted IDS is out of the lexicon. In addition, we report cross-domain recognition results on the public BCTR benchmark, which is used only to evaluate standard text recognition generalization in four domains.

\begin{table*}[h]
\centering
\small
\renewcommand{\arraystretch}{0.4}
\setlength{\tabcolsep}{8.5pt}
\caption{Ablation study of DTRNet. Starting from a text-only baseline, we gradually introduce the radical branch, query validity head, legality constraints, and IDS-guided confidence adjustment (IGCA). 
ACC and 1-NED are reported on \textit{test\_correct} and \textit{test\_faked}. 
Faked character detection is evaluated by Precision (P), Recall (R), and F1. 
IDS legality denotes the percentage of predicted IDS outputs satisfying the legality constraints.
}
\label{tab:ablation_main}
\begin{tabular}{c|c|cc|cc|ccc|c}
\toprule
\textbf{} & \multirow{2}{*}{\textbf{Method}}
& \multicolumn{2}{c|}{\textbf{test\_correct}}
& \multicolumn{2}{c|}{\textbf{test\_faked}}
& \multicolumn{3}{c|}{\textbf{Faked Char}}
& \multirow{2}{*}{\textbf{IDS legality}} \\
& 
& \textbf{ACC} & \textbf{1-NED}
& \textbf{ACC} & \textbf{1-NED}
& \textbf{P} & \textbf{R} & \textbf{F1}
& \\
\midrule

\multirow{3}{*}{\textit{Baseline}}
& Text-only                  & 84.58 & 96.10 & 0.34  & 74.79 & \textbf{54.32} & 0.86  & 1.69  & --    \\
& + Radical branch           & 84.98 & 96.20 & 19.61 & 80.11 & 44.24 & 34.87 & 39.04 & 77.56 \\
& + Query validity head      & 85.37 & 96.41 & 20.04 & \textbf{80.26} & 44.31 & \textbf{35.61} & \textbf{39.52} & 77.81 \\
\midrule

\multirow{2}{*}{\textit{Legality}}
& + Training penalty         & 85.60 & 96.43 & 19.97 & 80.36 & 44.31 & 35.16 & 39.21 & 77.73 \\
& + Inference constraint     & 85.82 & 96.46 & 20.29 & 80.25 & 45.05 & 34.53 & 39.10 & 87.03 \\
\midrule

\multirow{2}{*}{\textit{IGCA}}
& + Soft only                & 85.91 & 96.47 & 20.29 & 80.25 & 45.05 & 34.53 & 39.10 & 87.03 \\
& + Exact only               & 86.70 & 96.74 & 20.29 & 80.25 & 45.05 & 34.53 & 39.10 & 87.03 \\
\midrule

\textit{Full}
& DTRNet          & \textbf{86.81} & \textbf{96.76} & \textbf{20.29} & 80.25 & 45.05 & 34.53 & 39.10 & \textbf{87.03} \\
\bottomrule
\end{tabular}
\end{table*}

\subsection{Experimental Protocols}

\noindent\textbf{Implementation Details.} Models are trained and evaluated on the reconstructed Visual-C3 dataset. Images are resized to $32\times256$ with maximum text and IDS lengths capped at 15 and 25, respectively. The text vocabulary aligns with the 3,020-character lexicon, while the radical branch uses 323 radical and 12 structural tokens. We optimize using AdamW (LR = $6.5\times10^{-4}$, weight decay = 0.05) \cite{adamw_2017} for 100 epochs, employing PARSeqAug, a OneCycleLR scheduler (5-epoch warmup) \cite{CosineLR_2016}, and a batch size of 256 per GPU. The joint objective integrates text cross-entropy, IDS CTC, query validity, and legality penalty losses ($\lambda_{\text{text}} = \lambda_{\text{ids}} = \lambda_{\text{valid}} = 1.0$, $\lambda_{\text{illegal}} = 0.1$). During inference, faked characters are detected by matching the constrained beam-search IDS outputs against the lexicon, followed by IGCA to calibrate the text predictions. Following Visual-C3, we use the special symbol $X$ to denote faked characters. Experiments run with mixed precision on 2 NVIDIA A800 GPUs.

\noindent\textbf{Evaluation Metrics.} To evaluate performance, we design metrics across three dimensions. For basic text recognition, we report Line-level Accuracy (ACC)  and One Minus Normalized Edit Distance (1-NED) \cite{NED} on both the \textit{test\_correct} and \textit{test\_faked} subsets. Punctuation is normalized prior to evaluation.

For faked character detection, we formulate it as a character-level binary classification task targeting the special token $X$. To address potential length discrepancies, we align the predicted and ground-truth sequences globally via edit distance, and report character-level Precision, Recall, and F1 \cite{visual_c3} as the evaluation metrics.

Furthermore, to assess the structural stability in open-vocabulary scenarios, we introduce IDS Sequence Legality. Let $N_{\mathrm{seq}}$ be the total number of predicted text-line IDS sequences and $N_{\mathrm{legal}}$ be the number of structurally legal ones. The legality is computed as:
\begin{equation}
\mathrm{IDS\ Sequence\ Legality}=\frac{N_{\mathrm{legal}}}{N_{\mathrm{seq}}}.
\end{equation}
A sequence is counted as legal only when all valid character-level IDS predictions in the text line are legal. While not directly measuring detection accuracy, this metric reflects the structural reliability and validity of the IDS outputs used for verification.

\noindent\textbf{Adapting Methods for Faked Character Detection.}
We select 9 representative text recognition models as comparison baselines, including CRNN \cite{crnn_2017}, NRTR \cite{nrtr_2019}, ABINet \cite{abinet_2021}, SVTR \cite{svtr_2021}, PARSeq \cite{parseq_2022}, LISTER \cite{lister_2023}, SMTR \cite{smtr_2025}, CPPD \cite{CPPD_2025}, and SVTRv2 \cite{svtrv2_2025}. Since all these methods are designed to transcribe normal characters and do not support faked character detection, they need to be adapted to our unified protocol. Following the setting of Visual-C3, we apply a confidence-based rejection strategy to all baselines. Specifically, we first obtain the character-wise predictions and their confidence scores, and then replace any character whose confidence is below 0.2 with the special symbol $X$ to indicate a faked character. This adaptation does not modify the original recognition model itself, but only introduces a unified post-processing rule during inference.

\begin{figure}[h]
    \centering
    \includegraphics[width=0.68\columnwidth]{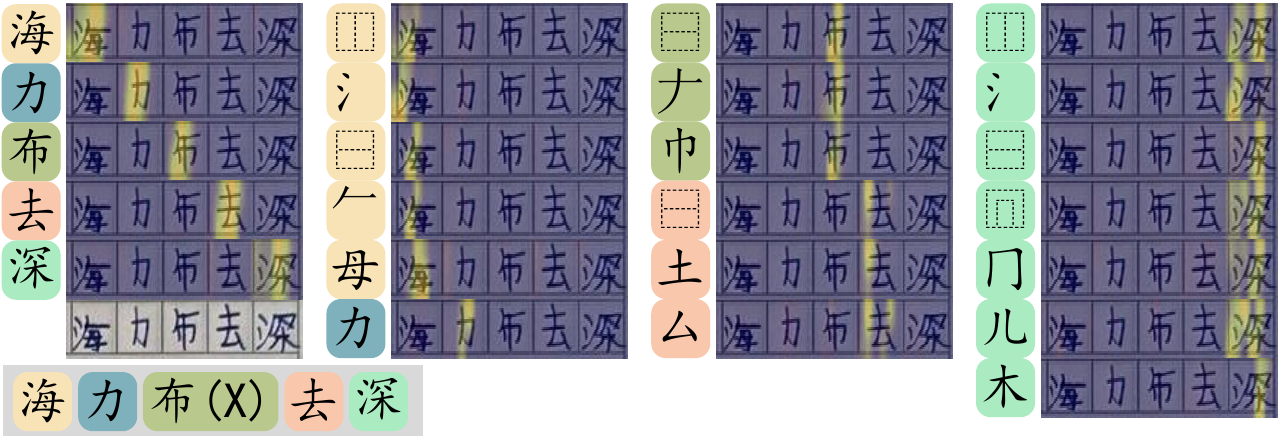}
    \caption{Visual analysis of faked characters in a line from \textit{test\_faked}. Left: character-wise localization attention maps showing the spatial focus of the structure branch. Right: predicted IDS components for each character. 
    }
    \Description{Character-wise localization and structural analysis for one handwritten line. Yellow attention bands show the spatial region assigned to each character, while adjacent colored columns list the predicted IDS components; the structurally abnormal character is output as X.}
    \label{fig:visual} 
\end{figure}

\subsection{Quantitative and Qualitative Results}
Table~\ref{tab:comparison_main} reports the baseline results under the default Visual-C3 confidence threshold of 0.2. The baseline recognizers degrade sharply on \textit{test\_faked}, even when they perform well on \textit{test\_correct}, indicating that they tend to over-correct faked characters into plausible normal ones. Under this protocol, DTRNet achieves the best overall performance, including 86.81\% ACC on \textit{test\_correct}, 20.29\% ACC and 80.25\% 1-NED on \textit{test\_faked}, and the best detection F1 of 39.10\%.

The detection gain mainly comes from recall: DTRNet reaches 34.53\%, exceeding CPPD by 12.28 points, while SVTRv2 shows that high precision alone is insufficient (66.67\% P but only 0.15\% R). DTRNet also produces highly legal IDS outputs (87.03\%). Fig.~\ref{fig:experiment_compare} shows the same trend: baseline methods usually normalize faked characters into legal ones, whereas DTRNet explicitly rejects them as $X$ with IDS evidence. These results indicate that explicit structural verification is more effective than confidence filtering for line-level faked character detection in handwritten Chinese text lines.

\subsection{Ablation Study}
\noindent\textbf{Structural Decoupling.} Table~\ref{tab:ablation_main} shows that the radical branch is essential for faked character detection. The text-only baseline almost fails on \textit{test\_faked} (0.34\% ACC and 1.69\% F1), whereas adding the radical branch raises F1 to 39.04\%. The query validity head further improves F1 to 39.52\%, indicating that explicit structural verification and stable localization are both important.

\noindent\textbf{Legality Constraints.} Legality modeling mainly improves the reliability of IDS outputs. The training penalty alone brings limited gains, while inference-time legality constraints increase IDS legality from 77.81\% to 87.03\% and slightly improve recognition, showing that grammar-aware decoding is more effective when enforced directly during inference under the line-level setting.

\noindent\textbf{IDS-Guided Confidence Adjustment.} IGCA mainly benefits normal text recognition rather than faked character judgment: Soft-only and Exact-only keep the detection metrics unchanged, but Exact-only improves \textit{test\_correct} ACC from 85.82\% to 86.70\%. The full DTRNet achieves the best overall result, confirming that the radical branch drives detection, legality constraints stabilize IDS prediction, and IGCA refines text recognition.

\subsection{Visual Analysis of Faked Characters}

To analyze how the radical branch handles faked characters, we visualize the token-level attention maps on a text line containing one faked character, as shown in Fig.~\ref{fig:visual}. A clear coarse-to-fine pattern can be observed: the structure token first attends to the whole character region, while the following component tokens progressively focus on smaller local parts. For legal characters, this process is stable and visually consistent, and the decoded token sequence can be mapped back to a valid entry in the character-IDS dictionary.

The faked character exhibits a different pattern. Its attention remains well localized on the target region, indicating that the failure is not caused by attention drift or region mismatch. Instead, the inconsistency appears after decomposition: although several local parts can be individually attended and decoded, their combination cannot form a valid character structure in the dictionary. This suggests that faked character detection is not a region-level recognition problem, but a structural validity checking problem based on the consistency of decomposed components.

\begin{table}[h]
\centering
\small
\caption{Comparison with OCR tools and MLLMs on the \textit{test\_faked} subset. We report recognition metrics (ACC and 1-NED) and faked character detection metrics (P/R/F1). The best result in each column is shown in bold.}
\label{tab:ocr_mllm}
\setlength{\tabcolsep}{6pt}
\renewcommand{\arraystretch}{0.4}
\begin{tabular}{c|cc|ccc}
\toprule
Method & ACC & 1-NED & P & R & F1 \\
\midrule
GPT-5.4 & 6.10 & 56.22 & 48.17 & 28.87 & 36.10 \\
qwen-vl-ocr & 0.10 & 73.33 & 50.00 & 0.16  & 0.36  \\
deepseek-v3.2 & 0.00 & 10.04 & 3.04  & 7.47  & 6.54  \\
 kimi-latest  & 3.10 & 67.82 & 52.39 & 15.65 & 24.10 \\
\midrule
PPOCRv5  & 1.30 & 73.25 & \textbf{69.38} & 2.86  & 5.49  \\
RapidOCR  & 0.30 & 72.81 & 61.90 & 1.09  & 2.15  \\
\midrule
DTRNet & \textbf{20.29} & \textbf{80.25} & 45.05 & \textbf{34.53} & \textbf{39.10} \\
\bottomrule
\end{tabular}
\end{table}

\subsection{Comparison with MLLMs and OCR-Tools}
Table~\ref{tab:ocr_mllm} shows that generic OCR tools and MLLMs remain weak on this task. DTRNet achieves the best results on \textit{test\_faked}, including 20.29\% ACC, 80.25\% 1-NED, and 39.10\% F1, while the strongest MLLM, GPT-5.4, reaches 36.10\% F1. OCR tools are conservative, yielding relatively high precision but extremely low recall, whereas MLLMs obtain slightly better recall but still tend to normalize faked characters into legal ones under semantic priors. This gap suggests that line-level faked character detection mainly depends on explicit character-structure verification, which is not explicitly modeled by general OCR tools or MLLMs under line-level handwritten settings.

\subsection{Standard Text Recognition Performance}
\label{experiments:bctr}

To further evaluate the generalization ability, we conduct additional experiments on the public \textit{Benchmarking Chinese Text Recognition (BCTR)}\cite{BCTR}, which spans four domains: \textit{Scene}, \textit{Web}, \textit{Doc}, and \textit{Handwriting (HW)}. Since BCTR is a standard text recognition benchmark rather than a faked character benchmark, we report only line-level accuracy to assess cross-domain robustness.

As shown in Table~\ref{tab:bctr_generalization}, our method achieves the highest average accuracy of 82.15\%, outperforming all compared methods. The gain is relatively small on \textit{Doc}, where most methods are already saturated, but more evident on the more challenging subsets. In particular, our method improves over SVTRv2 by 2.31\% on \textit{Scene}, 1.90\% on \textit{Web}, and 3.34\% on \textit{HW}. These results show that the proposed design generalizes well beyond the target benchmark and remains effective under substantial appearance and domain variations.

\begin{table}[h]
\centering
\small
\caption{Performance comparison on BCTR dataset. The best result in each column is shown in bold.}
\label{tab:bctr_generalization}
\setlength{\tabcolsep}{8pt}
\renewcommand{\arraystretch}{0.85}
\begin{tabular}{c|cccc|c}
\toprule
Method & \textit{Scene} & \textit{Web} & \textit{Doc} & \textit{HW} & \textit{Avg} \\
\midrule
CRNN    & 62.50 & 68.42 & 94.66 & 43.04 & 67.16 \\
NRTR    & 75.91 & 77.23 & 96.91 & 53.40 & 75.86 \\
ABINet  & 77.30 & 77.30 & 98.61 & 55.87 & 77.27 \\
LISTER  & 78.03 & 79.13 & 98.95 & 55.48 & 77.90 \\
CPPD    & 80.26 & 80.63 & 98.86 & 56.92 & 79.17 \\
SVTRv2  & 80.35 & 80.53 & 99.02 & 60.98 & 80.22 \\
\midrule
DTRNet  & \textbf{82.66} & \textbf{82.43} & \textbf{99.19} & \textbf{64.32} & \textbf{82.15} \\
\bottomrule
\end{tabular}
\end{table}

\subsection{Confidence Threshold Analysis}
\label{sec:threshold_analysis}

Although the confidence threshold of 0.2 used in Table~\ref{tab:comparison_main} follows the Visual-C3 protocol, different recognizers may exhibit different confidence calibration. We therefore conduct a model-wise confidence threshold analysis for CRNN, NRTR, CPPD, and SVTRv2. For each model, characters whose confidence falls below the threshold are replaced by $X$. We jointly examine the faked-character Char-F1 on \textit{test\_faked} and the line-level ACC on \textit{test\_correct}, so that improved rejection performance can be evaluated together with the false rejection of normal characters.
\begin{figure}[h]
    \centering
    \includegraphics[width=0.65\columnwidth]{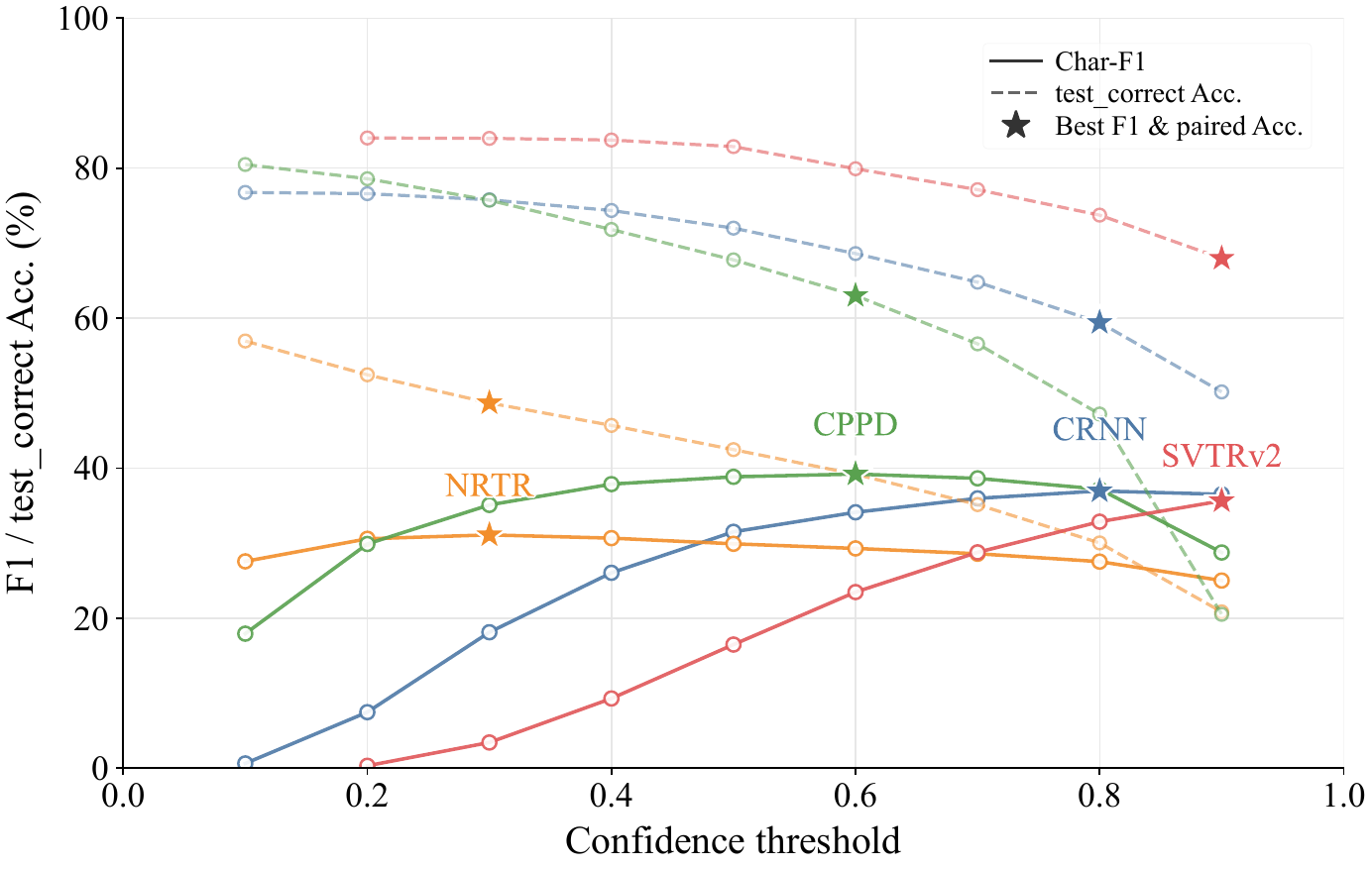}
    \caption{Confidence-threshold analysis of representative line-level recognition baselines. Solid curves denote faked-character Char-F1 on \textit{test\_faked}, dashed curves denote line-level ACC on \textit{test\_correct}, and stars mark the best-F1 operating point of each model.}
    \Description{Line chart comparing CRNN, NRTR, CPPD, and SVTRv2 across different confidence thresholds. Their faked-character F1 scores peak at different thresholds, while test-correct accuracy generally decreases as the threshold increases.}
    \label{fig:threshold_sweep}
\end{figure}

As shown in Fig.~\ref{fig:threshold_sweep}, the best confidence threshold varies substantially across models: 0.8 for CRNN but 0.3 for NRTR. Increasing the threshold generally improves faked-character detection until the model-specific optimum, but simultaneously reduces normal-text recognition accuracy. At its best-F1 operating point, CPPD reaches 39.21\% F1, while its paired \textit{test\_correct} ACC drops to 63.01\%. By comparison, DTRNet achieves a comparable F1 of 39.10\% while retaining 86.81\% ACC on normal text. These results show that confidence threshold tuning can improve the detection F1 of recognition baselines, but often at the cost of rejecting many normal characters. DTRNet instead provides a more favorable trade-off through explicit IDS-based structural verification.

\subsection{Limitations}
% \subsection{Limitations}
Despite the promising results, several limitations remain. 
First, line-level faked character detection is still a challenging problem, and there is still considerable room for improvement in detection accuracy, especially under complex handwriting patterns and highly ambiguous faked characters. 
Second, the current correction stage mainly relies on IGCA, which is a lightweight inference-time calibration strategy. Although effective, it does not fully exploit the potential of deeper interactions between text recognition and structural verification. More powerful structure-aware refinement or joint correction mechanisms deserve further investigation.

\section{Conclusion}

In this paper, we formulate line-level handwritten Chinese faked character detection as a generalized zero-shot problem and propose DTRNet, a dual Text-Radical decoding framework that combines context-aware text recognition with character-wise structural verification. By introducing IDS-based structural evidence and IDS-Guided Confidence Adjustment (IGCA), DTRNet can detect unseen faked characters while maintaining strong text recognition performance. Experiments on the reconstructed Visual-C3 benchmark show clear improvements over representative line-level baselines, OCR systems, and multimodal large language models, while results on BCTR further verify the robustness of the framework. 

\begin{acks}
This work is partially supported by grants from the National Natural Science Foundation of China under contract No. 62437001 and the Fundamental Research Funds for the Central Universities.
\end{acks}

\bibliographystyle{ACM-Reference-Format}
\bibliography{reference_final}

\newpage

% \title{Supplementary Material for DTRNet: Dual Text-Radical Decoding for Handwritten Chinese Text Recognition with Faked Character Detection}
\title[Supplementary Material for DTRNet]
{Supplementary Material for DTRNet: Dual Text-Radical Decoding for Handwritten Chinese Text Recognition with Faked Character Detection}

\makesupplementtitle

\renewcommand\thefigure{S\arabic{figure}}
\setcounter{figure}{0}

\renewcommand\thetable{S\arabic{table}}
\setcounter{table}{0}

\renewcommand\theequation{S\arabic{equation}}
\setcounter{equation}{0}

In the supplementary material, we provide the following.
\begin{itemize}
    \item In Section \ref{sec:task_benchmark}, we present the task, benchmark, and evaluation details, including the task setting, the reconstructed Visual-C3 benchmark, the Character-IDS lexicon details, and the evaluation metrics.
    \item In Section \ref{sec:quantitative}, we provide additional quantitative results, including comparison between CTC and autoregressive (AR) decoding in radical branch, extended analysis of the proposed design, and model complexity and efficiency analysis.
    \item In Section \ref{sec:qualitative}, we present additional qualitative results, including more faked character qualitative examples, qualitative comparisons with OCR tools and MLLMs, and qualitative results on BCTR.
    \item In Section \ref{sec:discussion}, we discuss the task setting, limitations of the current framework, and possible future directions.
\end{itemize}
\appendix

\section{Task, Benchmark and Evaluation Details}
\label{sec:task_benchmark}
\subsection{Task Setting and Reconstructed Visual-C3 Benchmark}
\label{sec:task_setting}

Our focus is \textbf{line-level faked character detection} rather than the full Visual-C3 task. Visual-C3 contains 10,072 samples, but only the annotated \textit{train} and \textit{val} splits (8,058 in total) are publicly available. The dataset contains both faked and misspelled characters with character-level annotations $(x,y,w,h,\mathrm{char})$. In this work, we keep only \textbf{faked character}-related samples and ignore misspellings, since misspellings are still valid characters and are closer to text-only CSC, whereas faked characters cannot be directly represented in plain text and require image and structural information.

\begin{figure*}[h]
    \centering \includegraphics[width=0.75\textwidth]{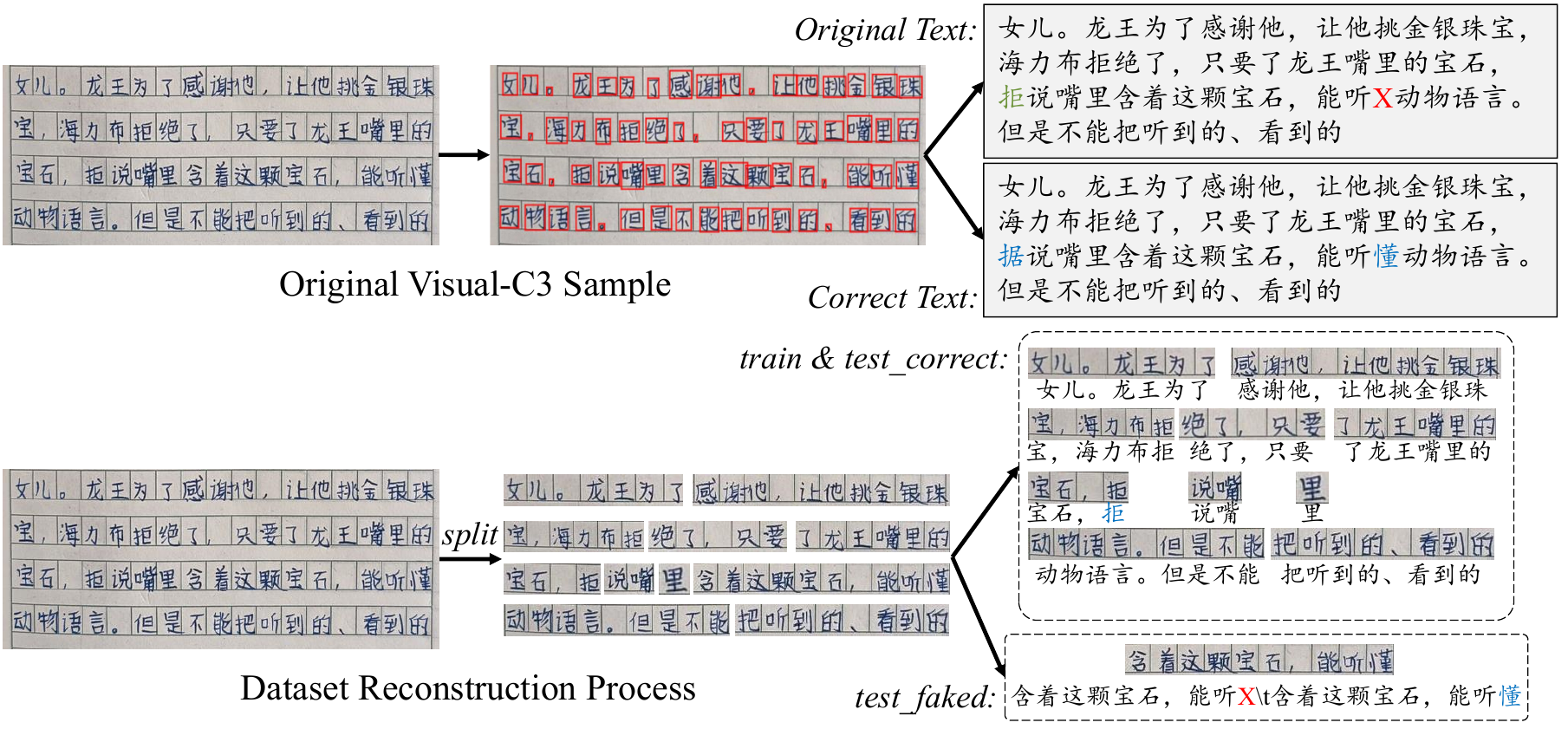} 
    \caption{Pipeline for reconstructing the original Visual-C3 dataset into our line-level faked character benchmark. In the labels, characters highlighted in red denote faked characters, while characters highlighted in green denote misspelled characters. We retain only faked character-related samples, use the original and corrected texts as paired labels, crop each multi-line image into single text lines, and further split them into short line segments. Normal segments are used to build the training set and \textit{test\_correct}, while segments containing faked characters are kept in \textit{test\_faked} for generalized zero-shot evaluation.}
    \Description{A left-to-right data pipeline from annotated Visual-C3 pages to line crops and short segments. Misspellings are discarded, normal segments form the training and test-correct splits, and segments with previously unseen faked characters form the test-faked split.}
    \label{fig:DatasetProcess}
\end{figure*}

Since faked characters in K-12 scenarios cannot be enumerated in advance, the task is inherently open-set. We therefore reconstruct Visual-C3 into a generalized zero-shot benchmark for line-level faked character detection. Specifically, we crop multi-line images into single text lines and further split them into short line segments. The training set contains only normal segments, while test set is divided into \textit{test\_correct} and \textit{test\_faked}. After reconstruction, the dataset grows from 8,058 to 59,189 samples, including 49,337 training samples, 5,482 \textit{test\_correct} samples, and 4,370 \textit{test\_faked} samples. All 5,119 faked characters are kept in \textit{test\_faked} and never appear in training. Figure~\ref{fig:DatasetProcess} shows the reconstruction pipeline. Accordingly, our line-level benchmark complements HCCEC evaluation on oracle character crops, covering text-line and isolated-character settings, respectively.
\subsection{Character-IDS Lexicon Details}
\label{sec:lexicon}

Instead of directly adopting an existing external decomposition dictionary, we build the Character-IDS lexicon to fully cover the reconstructed Visual-C3 benchmark. Specifically, the reconstructed dataset contains 3,020 target characters, and we use the same 3,020-character vocabulary in our text branch. This design is necessary because common Chinese decomposition resources do not fully cover all symbols appearing in the dataset, especially English letters, digits, and punctuation marks. Therefore, a lexicon built directly from the reconstructed benchmark is more suitable for our setting.

Among the 3,020 target characters, 2,843 have standard IDS mappings. For the remaining symbols without standard IDS decomposition, we do not discard them; instead, we treat them as atomic symbols and map them to themselves. This mainly includes English characters, digits, and Chinese or English punctuation, which allows the lexicon to achieve full coverage of the reconstructed Visual-C3 benchmark without introducing OOV symbols. Based on this construction, the radical branch uses 323 radical/atomic tokens and 12 structural tokens. The former includes both decomposable radicals and atomic symbols kept as themselves, while the latter corresponds to IDS structural operators. In addition, the original Visual-C3 labels do not contain the character $X$. Therefore, every $X$ used in this work is exclusively a special marker for faked characters and will not be confused with any real character class.

Unlike IDS-aware recognition methods that primarily use IDS to improve legal-character recognition, DTRNet uses predicted IDS as structural verification evidence. Accordingly, the 3,020-entry lexicon is specific to the current benchmark; rare or variant characters considered legal in other applications can be incorporated by adding their IDS or atomic mappings and extending the recognition vocabulary when needed.

\subsection{Evaluation Details}
\label{sec:eval}

For completeness, we provide the metric formulas omitted from the main paper. We evaluate text recognition with line-level Accuracy (ACC) and One Minus Normalized Edit Distance (1-NED) on both \textit{test\_correct} and \textit{test\_faked}. Faked-character detection is treated as a character-level binary classification task with $X$ as the positive label, using Precision, Recall, and F1. We also report IDS Sequence Legality to measure the structural validity of predicted IDS outputs. Punctuation is normalized before evaluation. Let $\hat{y}_i$ and $y_i$ denote the predicted and ground-truth sequences of the $i$-th sample, respectively, and let $N$ be the number of samples. Then
\begin{equation}
\mathrm{ACC}
=
\frac{\sum_{i=1}^{N}\mathbf{1}(\hat{y}_i=y_i)}{N}, \quad 
\mathrm{1\mbox{-}NED}
=
1-\frac{\sum_{i=1}^{N}\frac{\mathrm{EditDist}(\hat{y}_i,y_i)}{\max(|y_i|,1)}}{N}.
\end{equation}

For faked character detection, we first align the predicted and ground-truth sequences globally by edit distance, and then compute TP, FP, and FN on the special symbol $X$. The metrics are defined as
\begin{equation}
\mathrm{P}=\frac{\mathrm{TP}}{\mathrm{TP}+\mathrm{FP}}, \quad
\mathrm{R}=\frac{\mathrm{TP}}{\mathrm{TP}+\mathrm{FN}},\quad
\mathrm{F1}=\frac{2\mathrm{TP}}{2\mathrm{TP}+\mathrm{FP}+\mathrm{FN}}.
\end{equation}
Finally, let $N_{\mathrm{seq}}$ be the total number of predicted text-line IDS sequences and $N_{\mathrm{legal}}$ the number of structurally legal ones. IDS Sequence Legality is defined as
\begin{equation}
\mathrm{IDS\ Sequence\ Legality}
=
\frac{N_{\mathrm{legal}}}{N_{\mathrm{seq}}}.
\end{equation}
A text-line sample is counted as legal only when the IDS predictions of all valid character positions are legal. 

\section{Additional Quantitative Results}
\label{sec:quantitative}

\subsection{CTC vs.\ AR in the Radical Branch}
\label{sec:ctc_ar}

To further analyze the effect of different decoding strategies in the radical branch, we replace the original CTC with an autoregressive (AR). As shown in Table~\ref{tab:ctc_ar}, we compare two AR settings, namely \textit{AR\_wo\_IGCA} and \textit{AR}, with the full DTRNet. The two AR settings perform almost identically on \textit{test\_faked}, achieving only 1.44\% ACC, 75.29\% 1-NED, and 3.61\% F1. In contrast, DTRNet reaches 20.29\% ACC, 80.25\% 1-NED, and 39.10\% F1 on \textit{test\_faked}, indicating that CTC decoding is substantially more effective for faked character detection. On \textit{test\_correct}, the AR setting yields slightly better recognition results, suggesting that autoregressive decoding may still benefit normal character recognition.

A notable phenomenon is that the AR settings achieve extremely high IDS legality (99.84\%), far above the 87.03\% of DTRNet, while their detection recall remains only 1.85\%. This suggests that AR decoding tends to produce structurally legal IDS outputs that are biased toward common normal characters, thereby normalizing abnormal structures instead of preserving them for rejection. By contrast, the frame-wise independent transcription of CTC is more likely to retain such abnormal structures, making them less likely to match entries in the Character-IDS lexicon. These results are consistent with our design choice that the radical branch should prioritize faithful structural transcription over context-driven completion.

\begin{table*}[h]
\centering
\small
\renewcommand{\arraystretch}{0.9}
\setlength{\tabcolsep}{8.5pt}
\caption{Comparison between CTC and autoregressive (AR) decoding in the radical branch. 
ACC and 1-NED are reported on \textit{test\_correct} and \textit{test\_faked}. 
Faked character detection is evaluated by Precision (P), Recall (R), and F1. 
IDS legality denotes the percentage of predicted IDS outputs satisfying the legality constraints.
}
\label{tab:ctc_ar}
\begin{tabular}{c|c|cc|cc|ccc|c}
\toprule
\textbf{} & \multirow{2}{*}{\textbf{Method}}
& \multicolumn{2}{c|}{\textbf{test\_correct}}
& \multicolumn{2}{c|}{\textbf{test\_faked}}
& \multicolumn{3}{c|}{\textbf{Faked Char}}
& \multirow{2}{*}{\textbf{IDS legality}} \\
&
& \textbf{ACC} & \textbf{1-NED}
& \textbf{ACC} & \textbf{1-NED}
& \textbf{P} & \textbf{R} & \textbf{F1}
& \\
\midrule
\multirow{2}{*}{\textit{AR}}
& AR\_wo\_IGCA & 86.10 & 96.59 & 1.44 & 75.29 & 68.34 & 1.85 & 3.61 & 99.84 \\
& AR           & 86.93 & 96.94 & 1.44 & 75.29 & 68.34 & 1.85 & 3.61 & 99.84 \\
\midrule
\textit{Full}
& DTRNet       & 86.81 & 96.76 & 20.29 & 80.25 & 45.05 & 34.53 & 39.10 & 87.03 \\
\bottomrule
\end{tabular}
\end{table*}

\begin{table}[!h]
\centering
\small
\renewcommand{\arraystretch}{1}
\setlength{\tabcolsep}{5.5pt}
\caption{Model complexity and efficiency comparison with representative line-level methods, OCR tools, and MLLMs.
For line-level methods, Runtime is measured as model-only latency with batch size 1.
For OCR tools and MLLMs, Runtime is measured as average end-to-end wall-clock time per text line.
All runtime values are reported in ms/line.
F1 denotes the faked character detection result on \textit{test\_faked}.}
\label{tab:efficiency_all}
\begin{tabular}{c|c|c|c|c}
\toprule
\textbf{Type} & \textbf{Method} & \textbf{Params} & \textbf{Runtime} $\downarrow$ & \textbf{F1} $\uparrow$ \\
& & \textbf{(M)} & \textbf{(ms/line)} & \textbf{(\textit{test\_faked})} \\
\midrule
\multirow{6}{*}{Line-level}
& CRNN   & 17.69 & 7.28 & 7.46 \\
& ABINet & 42.86 & 13.96 & 1.91 \\
& CPPD   & 29.27 & 20.83 & 29.90 \\
& SVTRv2 & 20.94 & 17.73 & 0.31 \\
& Text-only & 22.48 & 19.86 & 1.69 \\
& DTRNet & 27.59 & 46.74 & 39.10 \\
\midrule
\multirow{2}{*}{OCR tool}
& PPOCRv5   & -- & 646.4  & 5.49 \\
& RapidOCR  & -- & 1474.9 & 2.15 \\
\midrule
\multirow{4}{*}{MLLM}
& GPT-5.4       & -- & 1843.4 & 36.10 \\
& qwen-vl-ocr   & -- & 608.6 & 0.36 \\
& kimi-latest   & -- & 832.7 & 24.10 \\
& deepseek-v3.2 & -- & 2470.7 & 6.54 \\
\bottomrule
\end{tabular}
\end{table}

\subsection{Extended Analysis of the Proposed Design}
\label{sec:extended_analysis}

We provide a more detailed analysis of the results in the main paper. As shown in Table 1 of the main paper, many baselines still achieve strong recognition accuracy on \textit{test\_correct}, but degrade substantially on \textit{test\_faked}. This suggests that good normal-text recognition does not naturally translate into effective faked character detection. In particular, methods relying mainly on contextual priors or confidence-based rejection tend to normalize faked characters into plausible legal ones. The advantage of DTRNet mainly comes from recall: it reaches 34.53\% recall, which is 12.28 points higher than CPPD, while SVTRv2, despite a precision of 66.67\%, only achieves 0.15\% recall. These results indicate that the key to this task is not merely higher recognition accuracy, but the ability to preserve and use structural evidence for abnormal characters.

Table 2 of the main paper also clarifies the role of each component. The radical branch is the main source of faked character detection capability: adding it to the text-only baseline increases F1 from 1.69\% to 39.04\%. The query validity head brings a further improvement, showing that more stable character localization benefits structural verification. Legality modeling mainly improves the reliability of IDS outputs, and the inference constraint is more effective than the training penalty, raising IDS legality from 77.81\% to 87.03\%. By contrast, IGCA mainly improves recognition on normal samples and has little effect on the detection metrics. Overall, the radical branch drives detection, legality constraints stabilize IDS prediction, and IGCA refines the text branch output.

A similar trend can be observed in Table 3 of the main paper. OCR tools usually show relatively high precision but extremely low recall, while MLLMs obtain slightly better recall but still tend to normalize faked characters into legal ones under semantic priors. GPT-5.4 is the strongest external model in our comparison, reaching 36.10\% F1, but it still remains below the 39.10\% of DTRNet. This further suggests that the main difficulty of line-level faked character detection lies not only in text-line transcription, but in judging character-level structural validity, which is not explicitly modeled by general OCR tools or MLLMs.

\subsection{Model Complexity and Efficiency Analysis}
\label{sec:efficiency}

Table~\ref{tab:efficiency_all} compares DTRNet with representative line-level methods, OCR tools, and MLLMs in complexity and runtime efficiency. Relative to the text-only baseline, DTRNet adds 5.11M parameters and increases the runtime from 19.86 to 46.74 ms/line, while improving F1 from 1.69\% to 39.10\%. The overhead mainly comes from character localization, IDS transcription, constrained decoding, lexicon matching, and IGCA. All line-level runtimes are measured with batch size 1 on the same NVIDIA A800 GPU.

Compared with OCR tools and MLLMs, DTRNet still runs substantially faster. As shown in the table, the per-line runtime of OCR tools and MLLMs generally ranges from hundreds to thousands of milliseconds, while their faked character detection performance remains unstable. Among them, GPT-5.4 gives the closest F1 to DTRNet, but its runtime is still much higher. Overall, DTRNet is not the fastest line-level model, but it achieves the best faked character detection result with acceptable computational cost, leading to a more favorable efficiency performance trade-off for this task.

\section{Additional Qualitative Results}
\label{sec:qualitative}

\subsection{More Faked Character Qualitative Examples}
\label{sec:more_qualitative}

\begin{figure*}[h]
    \centering
    \includegraphics[width=0.72\textwidth]{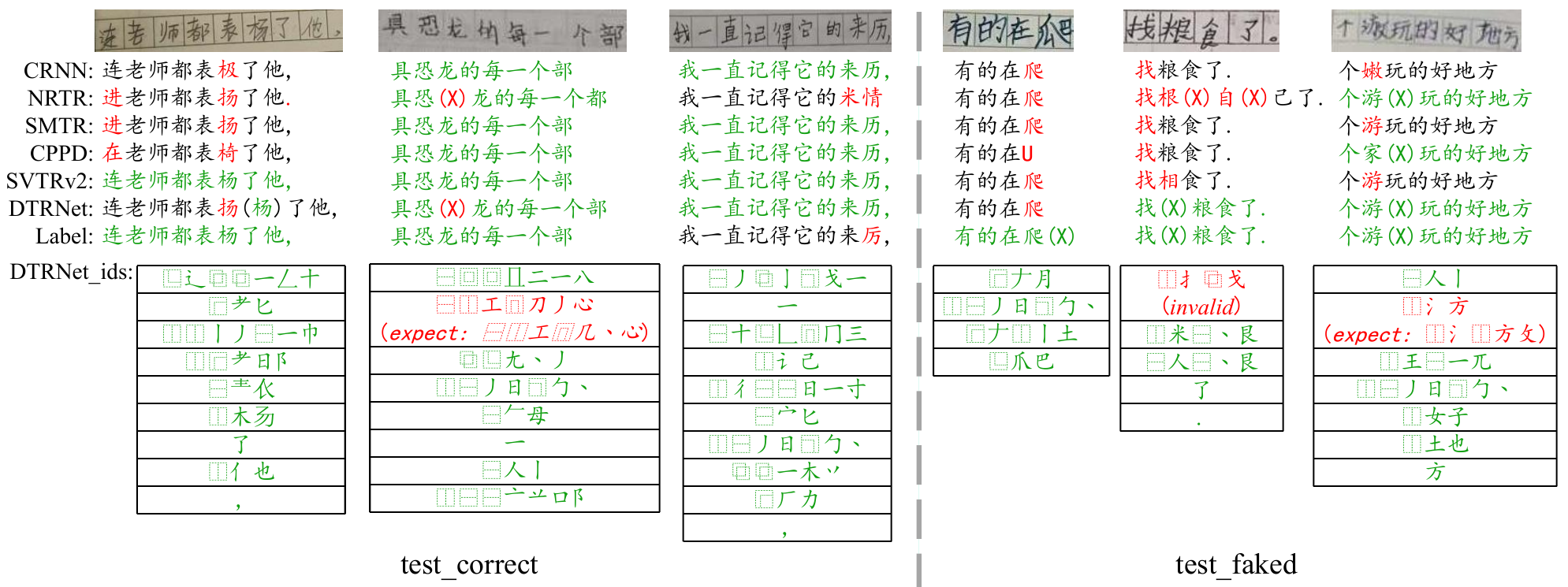}
    \caption{Additional qualitative failure cases of DTRNet on the reconstructed benchmark. The left three examples are from \textit{test\_correct}, and the right three are from \textit{test\_faked}. Green and red denote correct and incorrect predictions, respectively. Character-wise IDS outputs of DTRNet are also shown for diagnosis. The remaining errors mainly involve insufficient structural correction, false rejection, missed detection, and imperfect IDS decoding.}
    \Description{Six handwritten text-line failure cases with ground truth, DTRNet predictions, and character-wise IDS outputs. The examples illustrate insufficient structural correction, false rejection, missed faked characters, and invalid or inaccurate IDS decoding.}
    \label{fig:more_qualitative_failures}
\end{figure*}

We further present failure cases of DTRNet for qualitative analysis. As shown in Fig.~\ref{fig:more_qualitative_failures}, the left three examples are from \textit{test\_correct}, and the right three are from \textit{test\_faked}. In addition to the final text predictions, we also show the character-wise IDS outputs of DTRNet to analyze the interaction failures between the text and radical branches. On \textit{test\_correct}, the failures mainly arise from inconsistency between structural evidence and textual priors. In one case, the radical branch provides structural cues that are closer to the target character, but they are still insufficient to overturn the context-driven prediction of the text branch. In another case, a normal character is falsely rejected as $X$, indicating that ambiguous handwriting or unclear local structure may lead the radical branch to produce IDS outputs that deviate from legal matches. The figure also includes one sample with incorrect annotation in the original Visual-C3, where the apparent error is caused by the dataset annotation rather than the model itself.

On \textit{test\_faked}, the failures mainly appear as missed detections and insufficient IDS decoding. Missed detections indicate that some faked characters still remain structurally close to normal characters and are therefore absorbed into legal character classes. In other cases, the model correctly rejects the character as $X$, but the predicted IDS is still invalid or does not match the expected decomposition. These examples suggest that detecting a character as abnormal and accurately recovering its internal structure remain two different levels of capability in open-set faked character detection.

\begin{figure}[h]
    \centering
\includegraphics[width=1\columnwidth]{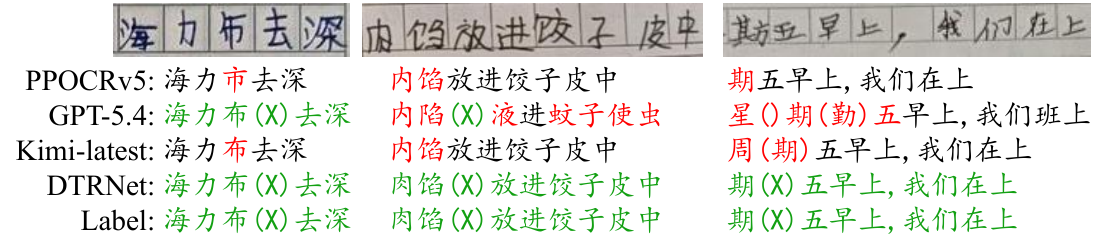}
    \caption{
    Qualitative comparisons with OCR tools and MLLMs on \textit{test\_faked}. We compare PPOCRv5, GPT-5.4, kimi-latest, and DTRNet. Green and red denote correct and incorrect predictions, respectively.}
    \Description{Three handwritten lines containing faked characters, with outputs from PPOCRv5, GPT-5.4, kimi-latest, and DTRNet. Green text marks correct predictions and red text marks incorrect predictions or over-corrections.}
    \label{fig:qual_compare_ocr_mllm}
\end{figure}

\subsection{Qualitative Comparisons with OCR Tools and MLLMs}
\label{sec:qual_compare}

\begingroup
\sloppy
We further present qualitative comparisons between DTRNet, OCR tools, and MLLMs on \textit{test\_faked}. For all experiments involving MLLMs, we use the prompt adapted from the Visual-C3 setting: {\footnotesize\texttt{First, we define the faked characters: a non-existent character in the dictionary resulting from the incorrect combination of radicals or components. Please recognize the provided handwritten image, and mark any faked character with an $X$. Return only a JSON object with exactly two string fields: \{"src":"...","tgt":"..."\}. The "src" field must preserve reading order and replace every faked character with a single uppercase $X$. The "tgt" field must preserve reading order, must not contain $X$, and should use the intended normal character at faked character positions. Do not output markdown, code fences, comments, or any extra keys.}}
\endgroup

As shown in Fig.~\ref{fig:qual_compare_ocr_mllm}, we compare PPOCRv5, GPT-5.4, kimi-latest, and DTRNet on \textit{test\_faked}. PPOCRv5 generally normalizes faked characters into plausible normal ones, and kimi-latest shows a similar tendency. GPT-5.4 exhibits stronger anomaly awareness and can correctly mark some faked characters as $X$, but in harder cases it also modifies neighboring normal characters, suggesting that its predictions still rely heavily on whole-line semantic generation. In contrast, DTRNet consistently rejects the faked character while preserving the remaining normal characters. These examples are consistent with the quantitative results and suggest that explicit character-wise IDS verification is important for avoiding over-correction of faked characters and reducing interference to surrounding context.

\subsection{Qualitative Results on BCTR}
\label{sec:bctr_qualitative}

\begin{figure}[h]
    \centering
    \includegraphics[width=1\columnwidth]{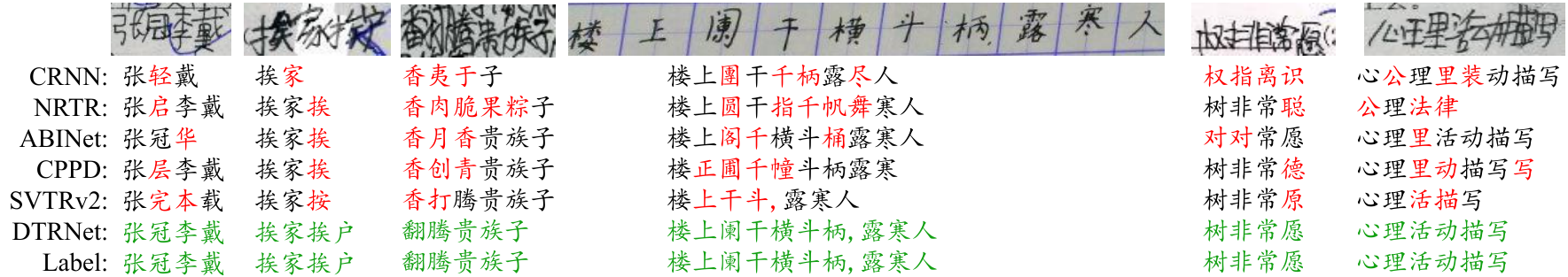}
    \caption{
    Qualitative results on the Handwriting (HW) subset of BCTR. We compare representative line-level methods and DTRNet. Green and red denote correct and incorrect predictions, respectively.}
    \Description{Six challenging BCTR handwriting samples with predictions from CRNN, NRTR, ABINet, CPPD, SVTRv2, and DTRNet. The samples contain background interference or tightly spaced characters; green denotes correct predictions and red denotes errors.}
    \label{fig:bctr_qualitative}
\end{figure}

We present qualitative results of DTRNet on the Handwriting (HW) subset of BCTR. As shown in Fig.~\ref{fig:bctr_qualitative}, we select six representative challenging examples and compare the outputs of CRNN, NRTR, ABINet, CPPD, SVTRv2, and DTRNet. These examples mainly involve two types of difficulty. One is strong background interference, such as correction marks, ruled backgrounds, or overlapping extra strokes. The other is extremely small spacing between adjacent characters, leading to unclear local boundaries. It can be seen that the errors of most baseline methods are concentrated in these locally high-interference regions, where they often fail to parse the overall structure of a character completely, resulting in partial recognition of a character, repeated recognition of local components, or omission of the whole character or some of its components. In contrast, DTRNet produces correct results on all these examples, indicating that character-wise structural modeling helps alleviate the recognition difficulties caused by complex backgrounds and tightly packed layouts. These cases are consistent with the quantitative results on BCTR in the main paper and further demonstrate that DTRNet also has strong robustness and generalization ability in standard handwritten text recognition.

\section{Discussion, Limitations, and Future Directions}
\label{sec:discussion}

\subsection{Discussion on the Task Setting}
\label{sec:task_discussion}

Different from standard HCTR, this work requires the model not only to transcribe handwritten text, but also to detect and mark faked characters in text lines. This setting is motivated by real K-12 educational scenarios. Students may write non-existent characters because of writing mistakes, memory lapses, or radical confusion. In such cases, teachers need to preserve these original errors for grading and feedback. If a system simply normalizes them into plausible legal characters, as in standard OCR, the original writing errors are lost. Therefore, the goal in this scenario is not only text transcription, but also explicit faked character detection.

We formulate this problem as a line-level generalized zero-shot task. In practice, handwriting recognition systems usually operate on text lines rather than isolated characters. Existing OCR pipelines first detect text regions at the line level and then directly recognize each line, instead of segmenting and recognizing characters one by one. Therefore, faked character detection in real applications should also be studied at the text-line level. This setting is also more efficient than character-level strategies such as TAN, since it avoids explicit character segmentation and repeated per-character processing. At the same time, faked characters are inherently open-set, since their categories cannot be enumerated in advance. We therefore reconstruct Visual-C3 into a line-level GZSL benchmark, where the training set contains only normal text lines and the test set contains both \textit{test\_correct} and \textit{test\_faked}. This provides a more realistic evaluation of generalization for line-level faked character detection.

\subsection{Limitations}
\label{sec:limitations}

Although DTRNet achieves strong results, line-level faked character detection remains far from solved. As shown by the failure cases in Sec.~\ref{sec:more_qualitative}, the remaining errors are not limited to missed detections, but also include inconsistency between the text and radical branches, false rejection of normal characters, and unstable IDS decoding. When a faked character is structurally close to a normal one, or when the handwriting itself is ambiguous, the model may capture some abnormal cues but still fail to reject it correctly. This suggests that the difficulty lies not only in anomaly detection itself, but also in the limited consistency among character localization, structural transcription, and text-structure interaction. In addition, although the current lexicon covers all benchmark characters, DTRNet depends on the completeness and correctness of application-specific mappings; missing or noisy mappings may cause legal characters to be falsely rejected, and this sensitivity has not yet been systematically evaluated.

Second, the current correction mechanism is still relatively lightweight. IGCA is essentially an inference-time calibration strategy. It uses structural evidence from the radical branch to alleviate over-correction caused by contextual priors, but its role is still mainly post-hoc adjustment rather than stronger joint modeling. In other words, the radical branch mainly serves as structural evidence and auxiliary correction, instead of forming a tighter bidirectional decision process with the text branch. As a result, when the two branches provide conflicting cues, the model still cannot stably perform structure-guided correction, which is also an important reason for the remaining branch inconsistency errors.

Finally, the current study is also limited by the available data resources. Visual-C3 provides an important foundation for this task, but its data form still differs noticeably from real K-12 scenarios. Most existing samples are grid-based composition-style handwriting, where characters are relatively well separated and the line layout is fairly regular. In real exams, however, faked characters may also appear in other question types and even other subjects, where handwritten text is often more tightly packed and accompanied by more complex page backgrounds and layout interference. Therefore, the current benchmark is more suitable for validating the method on relatively regular Chinese text, but still does not sufficiently cover broader K-12 scenarios. Future progress on this task will require larger, more diverse, and more realistic data resources.

\subsection{Future Directions}
\label{sec:future}

Based on the above limitations, future work can be pursued in several directions. First, at the model level, it is worth further strengthening structural modeling and cross-branch fusion. The current framework mainly models character structure at the radical level. In the future, it would be promising to explore finer-grained stroke-level modeling, or multi-granularity structural representations that combine radicals and strokes, so as to better capture the subtle differences between faked and normal characters. At the same time, the interaction between the text and radical branches is still relatively limited. How to align the two branches more accurately and produce a more stable final output on this basis remains an important problem for future study.

Second, at the task level, future work can extend the current setting from faked character detection to broader handwritten Chinese error analysis. In real K-12 scenarios, teachers need not only to mark faked characters, but also to annotate misspellings, redundant characters, missing characters, and even grammatical or semantic errors. In other words, the practical need is not simply to determine whether a character is abnormal, but to localize, identify, and annotate multiple types of errors in a unified manner. Therefore, combining faked character detection with broader error detection, localization, and correction is a more practically valuable direction.

Finally, at the data and evaluation level, future progress will also require larger, more diverse, and more realistic data resources for K-12 scenarios. Although the current benchmark already supports line-level open-set faked character detection, its data form is still relatively limited. Future datasets should cover more complex question types, layouts, and writing conditions, and should be paired with corresponding open-set evaluation protocols. This would enable a more systematic assessment of model generalization and diagnostic value in real educational scenarios.

\end{document}